\documentclass{vldb}
\usepackage{graphicx}
\usepackage{balance}  
\usepackage[]{algorithm2e}
\usepackage{multirow}

\begin{document}


\title{Generalization of metric classification algorithms for sequences classification and labelling}



%
%
%
%

\numberofauthors{3} 

\author{
%
%
\alignauthor
Roman Samarev\\
       \affaddr{Bauman Moscow State Technical University}\\
       \email{samarev@acm.org}
\alignauthor
Andrey Vasnetsov\\
       \affaddr{Bauman Moscow State Technical University}\\
       \email{vasnetsov93@gmail.com}
\alignauthor
Elizaveta Smelkova\\
       \affaddr{Bauman Moscow State Technical University}\\
       \email{lizuzu@mail.ru}
}


\date{14 Oct 2016}

\maketitle

\begin{abstract}

The article deals with the issue of modification of metric classification algorithms. In particular, it studies the algorithm k-Nearest Neighbours for its application to sequential data. A method of generalization of metric classification algorithms is proposed. As a part of it, there has been developed an algorithm for solving the problem of classification and labelling of sequential data. The advantages of the developed algorithm of classification in comparison with the existing one are also discussed in the article. There is a comparison of the effectiveness of the proposed algorithm with the algorithm of CRF in the task of chunking in the open data set CoNLL2000.

\end{abstract}

\section{Introduction}

The classification and labelling of sequences have a wide range of practical applications. In the research of the genome, classification of protein sequences in the existing categories is used to determine the function of a new protein\cite{kocsor2006application, sonego2008roc}. The study of the sequence of events to access Internet-based resources may allow distinguishing a user-person from a search bot\cite{feily2009survey}. The classification and labelling of sequences are used in Natural Language Processing (NLP)\cite{zhou2002named,lafferty2001conditional}. In General, a sequence is an ordered list of events $S = <s_1, s_2, \ldots, s_n>$ ,  where an event $s_i$, in general, can be associated with any data. However, in this work, we assume that the events are presented in the form of a feature vector of a fixed length, in which a feature can be represented by a string, the real number or integer, a boolean variable or a variable of an enumerated type. Each element of a sequence, among other things, may have a label; in this case, we speak of the labelled sequence.
Let L be a set of labels, then we have a function $Label:s_i \rightarrow l, l \in L$ which maps events to labels. In addition to the labels of the elements of a sequence, the sequence itself can also have the class: $Class: s_i \rightarrow c, c \in Cl$ where \textit{Cl} is the set of possible classes. 
Thus, when working with sequences, there are two tasks: a sequence classification — that is, the reproducing of the unknown function \textit{Class}, and a sequence labelling, that is, the reproducing of the unknown function \textit{Label}. . In the scientific literature\cite{nguyen2007comparisons}, this type of
classification is also called a strong sequence classification. Both tasks are supposed to have a
training set. It is a set of correctly classified or labelled sequences. A lot of different methods
have been proposed for solving the problem of classification. They can be divided into the
following groups:

\begin{itemize}
  \item A methods based on fixed size feature vectors. This method requires a conversion of a
sequence in a feature vector, and this conversion plays a major role in the
classification.

  \item A method based on an evaluation of the measure of distance between sequences. The
example of such a distance is Levenshtein distance\cite{levenshtein1966binary}.
  \item A classification using statistical models.
\end{itemize}

However, for the solutions of strong classification tasks, only the last group of methods is
suitable. The following ones should be mentioned\cite{nguyen2007comparisons}:

\begin{itemize} 
  \item Condition Random Fields, CRF.
  \item Hidden Markov Models.
  \item Markov SVM\cite{altun2003hidden}.
  \item Recurrent Neural Networks, RNN.
\end{itemize}

These methods are based on statistical modelling of any features of sequence elements and the
prediction of their labels based on these models. In some cases, this approach may be
ineffective, as, for example, despite the fact that the label signs have different categorical
values, they have a different degree of “proximity”. For example, the letters ``A'' and ``E'' in
some sense (as they sound) closer to each other than letters ``A'' and ``X''\cite{liu2006modified}, and the words
``red'' and ``blue'' are closer to each other than the words “algebra” and “stool”. It is often
convenient to set such proximity with the help of a distance function. To use such a distance
function in the classification process you need to have another type of algorithms, so-called
\textbf{metric classification algorithms}, for example, the algorithm k-Nearest Neighbours (kNN).
The correctness of the prediction of the element’s labels often depends on the context in
which this element is met. So, for example, the meaning of a word often depends on the
context in which it is used. But metric algorithms accept as input data a feature vector of a
fixed length and do not have an internal state, so their direct application to the problem of
labelling sequential data is not effective enough.
The article proposes a method of modifying metric classification algorithms that allows the
algorithms to consider the context of the element of a sequence, thereby improving the
classification accuracy.

\section{Generalization of classification algorithms for sequences}

To substantiate the possibility of modification of metric classification algorithms we consider
the relation
between the existing classification algorithms and methods applied to them to enable work
with sequences. One of the promising directions for solving strong classification is recurrent
neural networks (RNN), which are different from the usual neural networks in that fact that
they have feedback. This feedback means the connection between more logically remote
elements to less remote ones. The feedback connection allows storing and reproducing a
number of sequences of reactions to a single impulse. Due to the fact that the kNN algorithm
has no internal state, the use of generalization method which has been applied to the RNN is
not possible.

\begin{figure}
\centering
\includegraphics[width=0.9\linewidth]{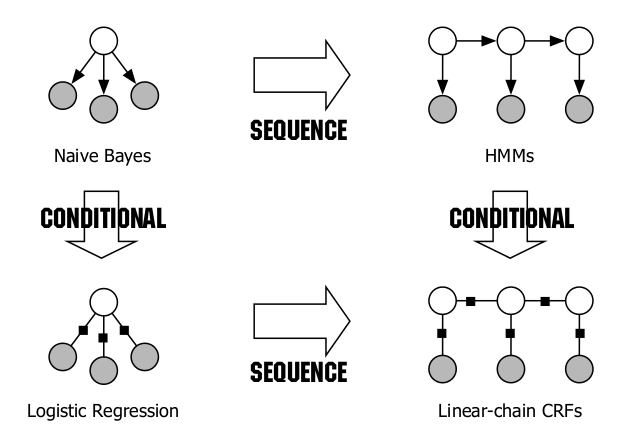}
\caption{Relations between classification algorithms}
\label{img:relation}
\end{figure}

Figure~\ref{img:relation} shows a diagram of the relationship between the algorithms of HMM, CRF, Naive Bayes and Logit Regression\cite{sutton2006introduction}. The diagram shows that the HMM algorithm is a
generalization of the naive Bayes classifier on the sequence of input data. Indeed, in the
working process of the HMM algorithm for each vertex, actually, the Bayes classification
algorithm with its own set of input data is applied. In this case for classification is used the
Viterbi algorithm, which finds a path in the graph that maximizes the probability of a
sequence.

Based on a review of the existing sequence classification algorithms we offer a method of
generalization of metric classification algorithms to deal with sequential data. The method
includes an algorithm for creating a graph model, the choice of the function being optimized,
the modified Viterbi algorithm. Let’s consider the proposed method on the example of the
kNN algorithm.
The kNN algorithm in the process of its work relies on the hypothesis of compactness: close
objects tend to lie in the same class, the close ones refer to the objects having the smallest
among all the other distances in certain, pre-determined metric. This hypothesis is naturally
generalized to the problem of sequence classification.

\begin{figure*}
   \centering
\includegraphics[width=0.9\linewidth]{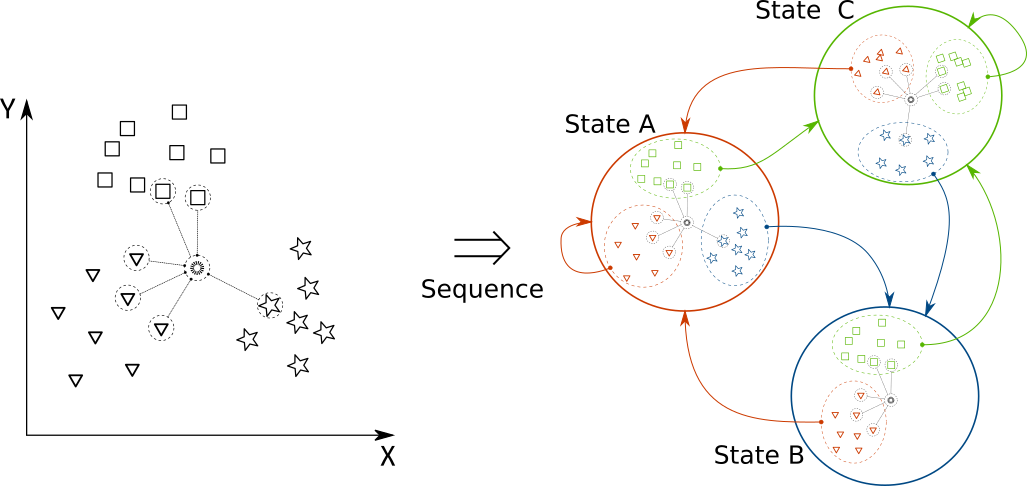}
   \caption{Structured k-Nearest Neighbours}\label{img:knn}
\end{figure*}

\textbf{The hypothesis of compactness of sequences}: each element of a sequence, as a rule, belongs
to the class which minimizes the total distance of all elements of this sequence to the known
elements of the appropriate class. Thus, the classification algorithm of sequences, which is a
generalization of kNN, is supposed to search for a sequence of classes which will minimize
the total distance of each element of the classified sequence to the nearest element of the
corresponding class in the training set. Since transitions between the classes in accordance
with the training set, form a kind of graph structure we will call the described algorithm as
Structured k-Nearest Neighbours (SkNN). The relationship of the algorithms is shown in
figure~\ref{img:knn}.

\subsection{Structured k-Nearest Neighbours definitions}
\label{sec:definitions}
To consider SkNN models we introduce the following definitions:
\begin{enumerate} 
  \item $k$ - a number of considered closest elements.
  \item $L$ - a set of possible labels of classes of features of sequences.
  \item $Cl$ - a set of possible classes of sequences.
  \item $S$ - a set of features of sequences.
  \item $X$ -  a set of sequences of a training set, $X = \{( s )_n : n \in \mathbb{N}, s \in S \}$
  \item $G = (V, E)$ - graph describing the structure of the classifier.
  \item $f_{vl}: V \mapsto L$ - function that determines the correspondence of vertex and labels. The function is injective.
  \item $f_{sl}: S \mapsto L$ - function determines the class label for each element of training set.
  \item $f_{sv}: S \mapsto V$ - function determines to which vertex the element of the training set
belongs to. All the first elements of sequences belong to $V_{init}$. 
  \item $C \in V$ - the current state of machine. At time $\tau = 0, C = V_{init}$.
\end{enumerate} 

The kNN algorithm is based on a directed graph $G = (V, E)$,  its each vertex corresponds to
either the beginning of the sequence (we call this vertex conventionally  $V_{init} \in V$),  or to a
possible class of feature of sequence $V_l, l \in L$, 
or to the end of the sequence $V_{end} \in V$.

The following conditions must be met:

\textbf{Condition 1.} Edge $E_{inti, x}$ is present in the graph $G$ if and only if the
training set has 2 consecutive elements $\alpha \in S$, and classes of elements $\beta \in S$,  $\alpha: L_{\alpha} = x$ and $\beta: L_{\beta} = y$.

\textbf{Condition 2.} Edge $E_{inti, x}$  is present if and only if a training set has a sequence the
first element of which has class $x$.

\textbf{Condition 3.} Edge $E_{x, end}$ is present if and only if a training set has a sequence the
last element of which has class $x$.

Creating the SkNN model from a training set is carried out with the help of the algorithm ~\ref{alg:const}.

\begin{algorithm}[h]
 \KwData{ Training set $X$ }
 \KwResult{ Constructed model of SkNN. }

 \For{$seq \in Dataset$}{
 	$C = V_{init}$\\
 	\For{$inst \in seq, inst \in S$}{
 		$l := f_{sl}(inst)$\\
    $L.add(l)$\\
 		$f_{sv}(inst) := C$;\\
 		$newC := f_{vl}(l)$;\\
 		$E.add( E_{C, newC } )$;\\
 		$C := newC$;\\
 	}
 	$E.add(E_{C, V_{end} })$;\\
 }

 \caption{Construction of SkNN model from training set.}
 \label{alg:const}
\end{algorithm}

For labeling sequences using the SkNN algorithm you can apply a modified Viterbi
algorithm, which is different from the original algorithm in the fact that it minimizes the total
distance. We give an example of Viterbi algorithm~\ref{alg:viterbi}.

\begin{algorithm}[h]
 \KwData{ Model of SkNN, unlabelled sequence $sequence$ }
 \KwResult{ Classified sequence. }

\For{$v_i \in V$}{
	$T1[i, 1] \leftarrow 0$\\
	$T2[i, 1] \leftarrow V_{init}$\\
}
\For{$inst_i \in sequence$}{
	\For{$v_j \in V$}{
		$T1[j, i] \leftarrow \underset{k}{min}( T1[k, i-1] + n\_dist(inst_i, {inst: f_{sv}(inst) = v_j}, K) )$ \\
		$T2[j, i] \leftarrow \underset{k}{argmin}( T1[k, i-1] + n\_dist(inst_i, {inst: f_{sv}(inst) = v_j}, K) ) $\\
	}
}
$T \leftarrow sizeof(sequence)$\\
$Z_T \leftarrow \underset{k}{argmin}(T1[k, T])$\\
$y_i \in Y$\\
\For{$i \leftarrow \{T, T-1, ..., 2\}$}{
	$Z_{i-1} \leftarrow T2[z_i, i]$\\
	$y_{i-1} \leftarrow V_{z_{i-1}}$\\
}

return $Y$

 \caption{Modified Viterbi algorithm.}
 \label{alg:viterbi}
\end{algorithm}

Function $n\_dist(inst, X, k)$ finds $k$ closest in some metric elements to $inst$ in the set $X$ and returns the weighted average distance to them.

\subsection{Types of features in SkNN}

In the current implementation of the model, as input for the algorithm SkNN, you can use
sequences of elements, each of which consists of a fixed number of features.
In the general case, the requirements for the type and composition of the characteristics of the
element are imposed solely by a distance function (metrics), which should be able to
determine the distance between any pair of elements.
It is a distinctive feature and advantage of the SkNN algorithm, as other algorithms such as
HMM, CRF, Structured SVM, RNN are able to work only with the description of the element
as a tuple of features of numeric and enumerated types.
For SkNN, however, it is possible to use any type of features, while the semantics of the
feature is not lost, but it can be represented as a distance function. For example, it is possible
to introduce metric which works with elements of the ontology. In this case, the distance
function can return, for example, the minimal path from one concept to the other, according
to the ontology. Using a weighted summation of different features, it is possible to create
metric that simultaneously takes into account the characteristics of arbitrary type, for
example, features derived from the ontology and the word vector, obtained by using the
algorithm word2vec\cite{mikolov2013distributed}.

\subsubsection{Obtaining graph structure using clustering}

In a large number of existing tasks of sequence classification it is required not to label each
individual element of a sequence but the entire sequence as a whole. In this case, input data
does not usually have label information of each single element and the SkNN algorithm is
reduced to a conventional kNN algorithm which is applied for each element separately.
It is evident that this approach does not give reasonable results, so it is necessary to mine
structure automatically in input data.
As the basis of SkNN is the comparison of elements in some metric, then it becomes possible
to use clustering for automatic mining of graph structure in sequences. To obtain such a
structure, you can perform the following actions:

\begin{itemize} 
  \item There is an initial structure in which each class of sequences is represented by a single
vertex. Features of sequences included in the corresponding class are associated with
vertex.
  \item For each vertex we hold clustering of related features of sequences.
  \item We split each clustered vertex into several vertices in such a way that each new vertex
corresponds to one cluster of old vertices and it contains features of the corresponding
cluster. In this case, edges are added in the graph in accordance with \textbf{condition 1}.
\end{itemize}  

It is worth noting that this approach is not necessary to use solely in conjunction with SkNN.
However, due to the fact that the input for clustering algorithms is a matrix of distances
between objects (or a distance function / similarity), the SkNN algorithm that uses the same
distance function will be native to clusterization.

\subsection{Sequence Classification}

As it has been mentioned above, in a large number of existing tasks of sequence classification
it is required not to classify each individual element of a sequence but the entire sequence as
a whole. SkNN’s Constructing Algorithm can be applied in this case without additional
modifications. You just need to perform the following actions:

\begin{enumerate} 
  \item For sequences of each class apply a clustering algorithm to obtain a graph
structure.
  \item Combine graphs obtained on the previous step. In order to have it:
      \begin{enumerate} 
      \item Add vertices \texttt{init} and \texttt{end}.
      \item Add edges from \texttt{init} to the initial vertices of the graphs, obtained in the first
step, and edges of the end vertices to the \texttt{end}.
      \end{enumerate}
\end{enumerate} 

\begin{figure}
\centering
\includegraphics[width=0.9\linewidth]{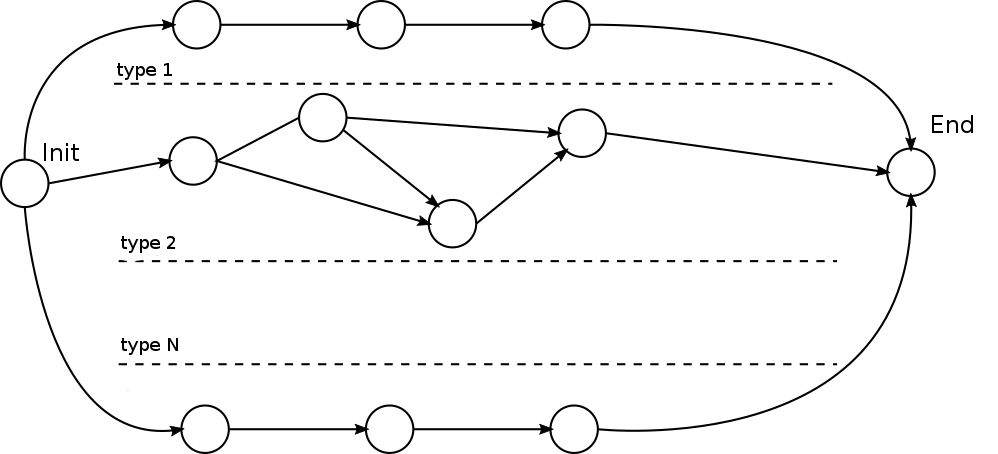}
\caption{General representation of sub-graphs for sequence classification}
\label{img:graph}
\end{figure}

The result is a graph consisting of non-overlapped sub-graphs. Example of general view of
this sub-graphs present in figure~\ref{img:graph}.

Applying the obtained graph in the SkNN algorithm for sequence classification, due to the
fact that in the graph there is no path from a sub-graph of one class to the sub-graph of a
different class, all the class labels will belong to the same sub-graph, and hence, one initial
class of sequence. Thus, for sequence classification, it is enough to determine which of the
initial classes the received labels belong to.

\section{Experiments}

To check the work of the algorithm and to compare it with the existing popular algorithms in
practice, we conducted a series of experiments using open datasets\cite{lichman2013uci}.
It should be noted that the field of possible application of the SkNN algorithm is quite big: as
standard benchmark datasets we used sets of such fields as recognition of handwritten
symbols (UJI Pen Characters \cite{llorens2008ujipenchars}) and processing of texts in natural language (CoNLL200\cite{tjong2000introduction}).

\subsection{Recognition of hand-written symbols}

For the first experiment we selected a dataset representing handwritten symbols. This dataset
was chosen because it contains a simple data type - a sequence of two-dimensional
coordinates on the plane. Simplicity of work with such data makes it easy to visualize, which
simplifies debugging of the algorithm. For testing the algorithm we compiled a training set
consisting of 200 sequences of writing handwritten digits of eight different authors. Test set
consisted of 40 sequences of 3 other authors. For the experiment we used the normalized
Euclidean distance, as the simplest example of a metric. The experiment showed that the
accuracy of the SkNN is equal to 92.5\%, which is significantly higher than random guessing
(10\% in this case).

It is worth noting, even though the accuracy of the algorithm SkNN does not exceed the
accuracy of the algorithms, which are intended for solving this very problem\cite{ramos2007input}, but it
shows the result that can be compared with them.

\subsection{Chunking}

The second experimental dataset \textbf{CoNLL200} represents sequences of words in sentences in
English with the corresponding parts of speech derived (POS-tag) with the help of program
Brill tagger\cite{acedanski2010morphosyntactic} and chunk tag which has been obtained from corpora WSJ\cite{paul1992design}. Group labels
represent such parts of the sentence as verb groups, e.t.c..
In the experiment the accuracy of the SkNN algorithm was compared with the accuracy of
the CRF algorithm implemented in the software utility CRF++ . This algorithm is one of the
most popular and accurate algorithms today in the field of NLP.
Dataset parameters: the size of training corpora — 9 thousand sentences, 220 thousand
words, the size of the test set - 400 sentences, 10 thousand words. As a hyperparameter for
both algorithms there is a size of context window which is added to the current feature of
sequence. The accuracy of the algorithms was compared with equal values of this
hyperparameter.

Another parameter for the SkNN algorithm is metric which is used for distance calculation.
The experiment was carried out using different metrics which were most successfully applied
in practice in the field NLP\cite{daelemans2005memory}. Metrics which have been used:

\begin{enumerate} 
  \item Modified Value Difference Metric (MVDM) \cite{liu2006modified}.
  \item Overlap.
  \item Weighted Overlap.
    \begin{enumerate}
      \item Using Information Gain\cite{kent1983information}.  
      \item Using Information Gain Ratio\cite{mori2002information}.
    \end{enumerate}
\end{enumerate} 

The results of the experiments are shown in table~\ref{table:results}.

\begin{table}
  \centering
  \caption{Experiment results}
  \label{table:results}
  \begin{tabular}{|l|l|l|l|}
    \hline
    \textbf{\begin{tabular}[c]{@{}l@{}}Context \\ window
    \end{tabular}} & \textbf{Algorithm} & \textbf{Metric} & \textbf{Accuracy} \\ \hline
    \multirow{4}{*}{\begin{tabular}[c]{@{}l@{}}Current \\ element only\end{tabular}} & SkNN & MVDM & 0.71\\ \cline{2-4} 
    & SkNN & Overlap & 0.7 \\ \cline{2-4} 
    & SkNN & \begin{tabular}[c]{@{}l@{}}IG \\ Weighted\\ Overlap\end{tabular} & 0.76 \\ \cline{2-4}  
    & CRF & - & 0.85 \\ \hline
    \multirow{5}{*}{\begin{tabular}[c]{@{}l@{}}Two elements \\ before and after \\ current one.\end{tabular}} & SkNN & Overlap & 0.91 \\ \cline{2-4} & SkNN & \begin{tabular}[c]{@{}l@{}}IG\\ Weighted \\ Overlap\end{tabular} & 0.6 \\ \cline{2-4} 
    & SkNN & \begin{tabular}[c]{@{}l@{}}IGRatio\\ Weighted \\ Overlap\end{tabular} & 0.88 \\ \cline{2-4} 
    & \textbf{SkNN} & \textbf{MVDM} & \textbf{0.93} \\ \cline{2-4} 
    & CRF & - & 0.87 \\ \hline
  \end{tabular}
\end{table}

\section{Conclusions}
This article studied the possibility of modification of metric classification algorithms on the
example of the algorithm k-Nearest Neighbours for their application to sequences. The
scientific novelty lies in the fact that new method of generalization of metric classification
algorithm is proposed. With using of this method new algorithm for sequence classification
SkNN was offered and it demonstrated its effectiveness in problems of classification and
labelling in the field of NLP. A comparison of the algorithm with the other often used
algorithms shows that with the accuracy of the algorithm it is reasonable to continue working
towards its effective realization and, for example, the implementation of this algorithm for
computing clusters, which will significantly increase its performance.


\balance

\bibliographystyle{abbrv}

\bibliography{ref} 


\end{document}